\documentclass[letterpaper, 10 pt, conference]{ieeeconf}  

\IEEEoverridecommandlockouts                              

\usepackage{graphics} 
\usepackage{epsfig} 
\usepackage{mathptmx} 
\usepackage{times} 
\usepackage{amsmath} 
\usepackage{amssymb}  
\usepackage{xcolor}
\usepackage{url}

\title{\LARGE \bf Design and implementation of a parsimonious neuromorphic PID \\for onboard altitude control for MAVs using neuromorphic processors}

\author{Stein Stroobants, Julien Dupeyroux and Guido C.H.E. de Croon
\thanks{*This work has received funding from the ECSEL Joint Undertaking (JU) under grant agreement No. 826610. The JU receives support from the European Union's Horizon 2020 research and innovation program and Spain, Austria, Belgium, Czech Republic, France, Italy, Latvia, Netherlands. It has also received funding from the European Office of Aerospace Research and Development (EOARD).}
\thanks{All authors are with the Micro Air Vehicle Lab, Faculty of Aerospace
Engineering, Delft University of Technology, The Netherlands. Contact: 
        {\tt\small s.stroobants@tudelft.nl}}%
}

\begin{document}

\maketitle
\thispagestyle{empty}
\pagestyle{empty}

\begin{abstract}

The great promises of neuromorphic sensing and processing for robotics have led researchers and engineers to investigate novel models for robust and reliable control of autonomous robots (navigation, obstacle detection and avoidance, etc.), especially for quadrotors in challenging contexts such as drone racing and aggressive maneuvers. Using spiking neural networks, these models can be run on neuromorphic hardware to benefit from outstanding update rates and high energy efficiency. Yet, low-level controllers are often neglected and remain outside of the neuromorphic loop. Designing low-level neuromorphic controllers is crucial to remove the standard PID, and therefore benefit from all the advantages of closing the neuromorphic loop. In this paper, we propose a parsimonious and adjustable neuromorphic PID controller, endowed with a minimal number of 93 neurons sparsely connected to achieve autonomous, onboard altitude control of a quadrotor equipped with Intel's Loihi neuromorphic chip. We successfully demonstrate the robustness of our proposed network in a set of experiments where the quadrotor is requested to reach a target altitude from take-off. Our results confirm the suitability of such low-level neuromorphic controllers, ultimately with a very high update frequency.
\end{abstract}

\section{Introduction}

In the coming years, autonomous drones are expected to perform a wide range of complex tasks in unknown environments. These tasks include autonomous take-off and landing, dynamic obstacle detection and avoidance, long-range navigation, etc~\cite{floreano2015science}. In spite of all the major accomplishments over the last decades in aerial robotics research, drones still cannot compete with their biological counterparts such as flying insects and birds. Indeed, these animals perform similar tasks using less computational power at a higher energy efficiency while being more robust to disturbances like wind gusts. For instance, despite their mere 100,000 neurons, fruit flies nimbly perform aggressive flight maneuvers and chase for mates while avoiding obstacles in complex, cluttered environments~\cite{zheng2018complete}. Similarly, desert ants \textit{Cataglyphis} are indisputably champions at long-range navigation in the desert, yet they only have 250,000 neurons to ensure such impressive performance~\cite{muller1988path}. In contrast, state-of-the-art quadrotors are often equipped with heavy, energy-consuming computing units like Graphics Processing Units (GPU) to enable Artificial Intelligence (AI) based solutions for in-flight autonomy such as vision-based obstacle avoidance in racing tasks~\cite{jung2018perception, kaufmann2019beauty}. As a matter of fact, the application of conventional neural networks for Micro Air Vehicles (MAVs) is limited by the energy consumption, weight and synchronous nature of the available hardware. This is particularly true in the context of vision-based control where the use of deep Convolutional Neural Networks (CNNs) severely hampers MAVs' flight duration. In addition, the forecast of the end of Moore's law in the coming decades suggests that the pressure on embedded processing will increase with drones' autonomy~\cite{lundstrom2003moore, theis2017end}. 

This analysis has driven researchers to put efforts in developing novel forms of information representation and processing, such as asynchronous Spiking Neural Networks (SNNs) to more closely model natural neurons and synapses and benefit from their overall performance~\cite{mahowald1991silicon, ghosh2009spiking}. These networks offer ample opportunity for a higher energy efficiency and faster computation but require dedicated neuromorphic hardware that can deal with the analog nature of the neuron membrane dynamics. Over the last decade, efforts have been made to develop the first neuromorphic processors to run SNNs, such as HICANN~\cite{schemmel2010wafer}, NeuroGrid~\cite{benjamin2014neurogrid}, IBM's TrueNorth~\cite{merolla2014million}, APT's SpiNNaker~\cite{furber2014spinnaker} and Intel's Loihi~\cite{davies2018loihi}. This new generation of processors yields great opportunities for the application of SNNs in aerial robotics, especially for MAVs where energy, payload and processing time are crucial. Autonomous control for quadrotors requires low latency and high computational power at low energy cost. Such performance can be achieved by means of neuromorphic algorithms like SNNs running on neuromorphic hardware. 

In previous work, we introduced an evolved SNN to control the landing of a MAV equipped with the Loihi chip~\cite{dupeyroux2020neuromorphic}. Made of only 35 spiking neurons, this model used the divergence of the ventral optic flow to determine the thrust set-point to send to a low-level PID controller running on the von Neuman CPU. In order to make the whole processing neuromorphic and therefore increase the control loop frequency, neuromorphic low-level controllers are required. Examples of such systems have been proposed and include controllers for open- and closed-loop controllers for DC motors~\cite{fernandez2012motorpid, pena2013pidmotoropenloop}, robot manipulators~\cite{dewolf2016armcontrol}, optic-flow based landing and neuromorphic implementations of standard PID controllers~\cite{zaidel2021neuropid, stagsted2020towards}. The computational and logical simplicity of PID controllers makes them an interesting choice for many control tasks. 

In Stagsted et al.~\cite{stagsted2020towards}, the mathematical operations of the conventional PID were implemented in a position-coded SNN by utilizing \textit{operation} arrays. Using such a position-coding, the precision of the controller corresponds to the number of neurons used for the encoding. To obtain an error precision of $N$, these arrays contain $N \times N$ neurons. Combining these arrays, a controller for a 1-DOF birotor fixed to a frame was designed and executed on Intel's Loihi chip to control the roll angle by inputting measurements of both the angle and the angular velocity. However, the influence of the derivative term in the response of the controller was not as important as for the proportional and integral terms, thus explaining the oscillations observed at zero-roll angles. Besides, the amount of necessary neurons for one single PID controller was restricting the precision of the representation of error and command values. To allow for smooth control the change in output command should have a large precision. Autonomous control with a cascaded-PID controller of a MAV in mid-air requires a bare-minimum of 6 PID controllers (but preferably more to include velocity control as well). Alternatively, in~\cite{zaidel2021neuropid}, a neuromorphic PID was designed to calculate the three terms of the controller using rate-coded signals, and further tested on-board the Loihi chip. Interestingly, rate-coding allowed to improve the derivative term by means of a set of different time scales for the synapses. However, such a solution remains hard to implement on the Loihi as the chip currently does not support high synaptic time constants.

These pioneer studies demonstrated the feasibility of using neuromorphic PIDs to control robots using neuromorphic processors such as the Loihi. Yet, having such systems working online to control a free-flying robot remains a great challenge, especially in the context of aerial vehicles where the robustness, reliability and execution speed of controllers is crucial. In this work, a different method of performing the mathematical adding (and subtracting) operations is suggested that reduces the number of necessary neurons by an order of magnitude as compared to the proposed method in~\cite{stagsted2020towards}, while also allowing for a non-linear distribution of the position-coded spikes that represent both the input, error and output floating point values. This simultaneously allows for full control of a MAV with a higher precision while keeping a smaller distribution, and thus more precision of the control output around zero. Furthermore, we successfully demonstrate the performance of our neuromorphic PID controller running on-board the Loihi neuromorphic chip for the altitude control of a free-flying MAV.

\section{Methods}

In the following, we introduce the standard cascaded PID controller and the neuromorphic PID (N-PID) for a quadrotor. The cascaded PID is hereafter considered as the baseline for comparison with the neuromorphic PID. 

\subsection{Cascaded PID controller}
\begin{figure}
    \centering
    \includegraphics[width=\linewidth]{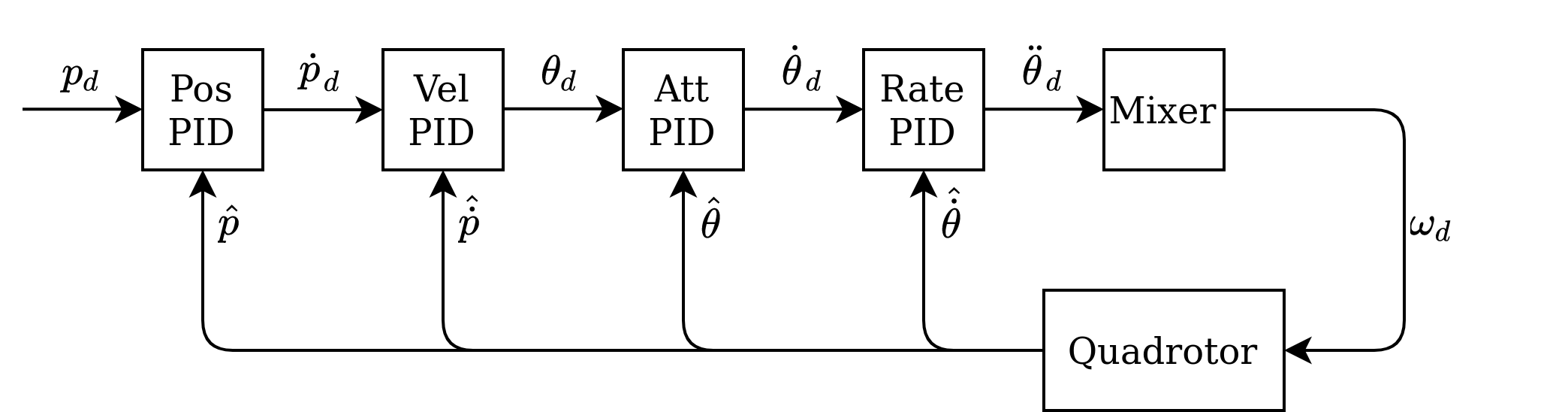}
    \caption{Cascaded PID for multirotor control where $p$ is the position ($x$, $y$, and $z$) and $\theta$ are attitudes ($\theta$, $\phi$, and $\psi$), a dot ($\dot{a}$) represents the time-derivative whereas a hat ($\hat{a}$) stands for the measurement. $\omega_d$ are desired rotor speeds, calculated in the control mixer from the torque/thrust values.}
    \label{fig:cascadedpid}
\end{figure}

The cascaded PID controller used in quadrotors consists of multiple interconnected PIDs that produce rotor-speed commands for all four rotors. The design of the controller can be seen in Fig.~\ref{fig:cascadedpid}, where each PID block represents a set of multiple PID controllers running in parallel to process the commands for the three axes ($x$, $y$, $z$). The continuous-time definition of a given PID controller is provided by the following set of equations: 

\begin{equation}
    \left\{
    \begin{aligned}
     u(t) &= K_P \Big( e(t) + \frac{1}{T_I} \int_0^t e(\tau) d\tau + T_D \frac{de(t)}{dt}\Big) \\
    e(t) &= r(t) - y(t)
    \end{aligned} \right.
\end{equation}

\noindent where $u(t)$ is the control signal at time $t$, $K_P$ the proportional gain, $T_I$ and $T_D$ the integral and derivative time-constants respectively and $e(t)$ the error signal between the target $r(t)$ and the measurement $y(t)$. This can be transformed into a discrete-time PID controller where the signals are defined as follows:

\begin{equation}
    \left\{
    \begin{aligned}
        e_k &= r_k - y_k \\
        i_k &= i_{k-1} + e_k \Delta t \\
        d_k &= \frac{e_k - e_{k-1}}{\Delta t} \\
        u_k &= K_P e_k + \frac{K_P}{T_I} i_k + K_P T_D d_k
    \end{aligned} \right.
\end{equation}

As illustrated in Fig.~\ref{fig:cascadedpid}, the inputs to the first layer of the controller are position set-points, which produce velocity set-points and so on. The outputs of the last PID are the thrust and torque commands that are translated into rotor speed commands in the control mixer based on the model parameters of the quadrotor.

\subsection{Neuromorphic PID}

The neuromorphic PID, referred to as the N-PID, is achieved by using populations of neurons to perform mathematical operations (i.e., additions and subtractions) necessary for computing the output command of the controller.

\subsubsection{Input neurons population}

We propose to use a set of~$N$~input neurons to encode the floating-point values using a standard position-coding scheme: each neuron will fire whenever the input value falls within a predefined range. The distribution of those sensitivity ranges can either be uniform, meaning that all neurons will have the same sensitivity, or non-uniform. In the latest, for instance, a quadratic distribution can be used to increase the neurons' sensitivity around a certain point of interest. Using such non-uniform, arbitrary distributions opens the opportunity for more accurate control around a certain set-point. To encode a floating-point value as a position-coded value, the distance from the floating-point value to all the $N$ values represented by the encoded distribution is calculated. The neuron for giving the lowest error will then fire, while the other neurons will stay inactive (\textit{winner-takes-all}).

\subsubsection{Aggregate population layer}

\begin{figure}
    \centering
    \includegraphics[width=0.82\linewidth]{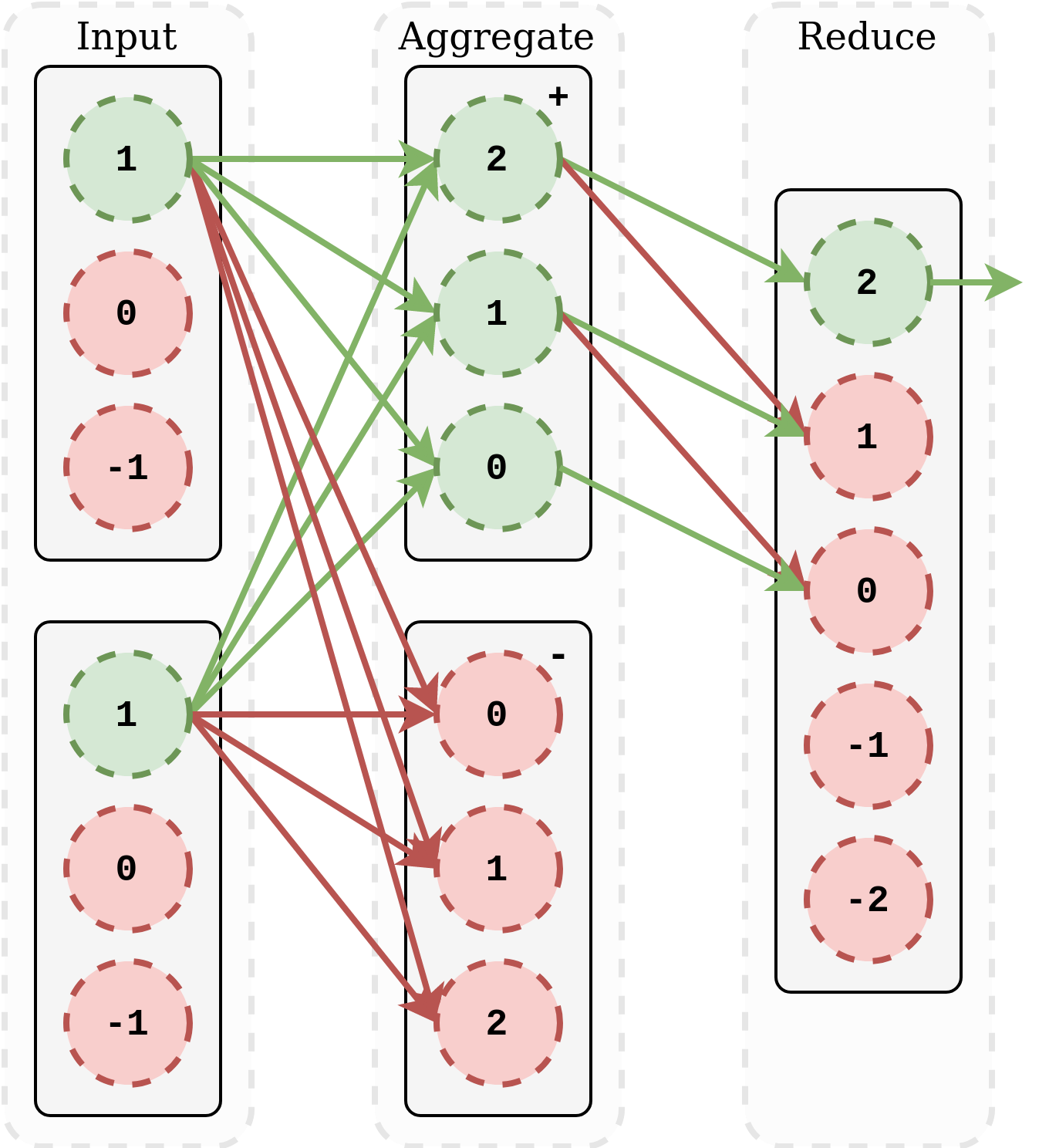}
    \caption{Adder neuron population for two inputs of [-1, 0, 1] and an output of [-2, -1, 0, 1, 2]. This example shows the addition of the values 1 and 1. The green and red arrows represent inhibitory and excitatory spikes respectively. The membrane potential of all neurons in the positive group in the aggregate layer sums to 2, and in the negative group to -2. All neurons in the positive group (with thresholds indicated in the figure) will now fire and in the reduce layer only the neuron representing the value 2 will fire. }
    \label{fig:spiking_adder}
\end{figure}

In the proposed N-PID, a population of neurons performs either an addition or a subtraction with two layers (Fig.~\ref{fig:spiking_adder}): an \textit{aggregate} layer performing the mathematical operation, then followed by a \textit{reduce} layer that implements a position-coded winner-takes-all. The aggregate layer consists of two sub-populations of neurons, one for positive and one for negative outputs . The aggregate layer is densely connected to a set of input layers. For instance, if we aim at adding two input signals, there will be two populations of input neurons as represented in Fig.~\ref{fig:spiking_adder}. To ensure the mathematical operation, the corresponding synaptic weights are linearly proportional inputs. However, connections to the negative sub-layer in the aggregate layer are multiplied by~$-1$. This way, the negative values are represented by their absolute and the use of negative thresholds is bypassed. The thresholds of the aggregate neurons are proportional to the absolute of the values in the output range. This way, all neurons with a threshold lower than the sum of the input values in the correct sub-population will fire.

\subsubsection{Reduce population layer}

The connection between neurons in the aggregate layer and neurons in the reduce layer is designed as follows (Fig.~\ref{fig:spiking_adder}): first, an excitatory synapse connects the two neurons representing the same value, and second, an inhibitory synapse connects the same input neuron (aggregate) to the adjacent neuron in the reduce layer. Only the highest firing neuron in the aggregate layer will spike in the reduce layer. Changing from addition to subtraction simply requires flipping the sign of the weights of the input to be subtracted. This brings the total amount of neurons necessary for performing an addition $2N + 1$ where $N$ is the resolution.

\subsubsection{Combining into a single N-PID unit}

\begin{figure}
    \centering
    \includegraphics[width=\linewidth]{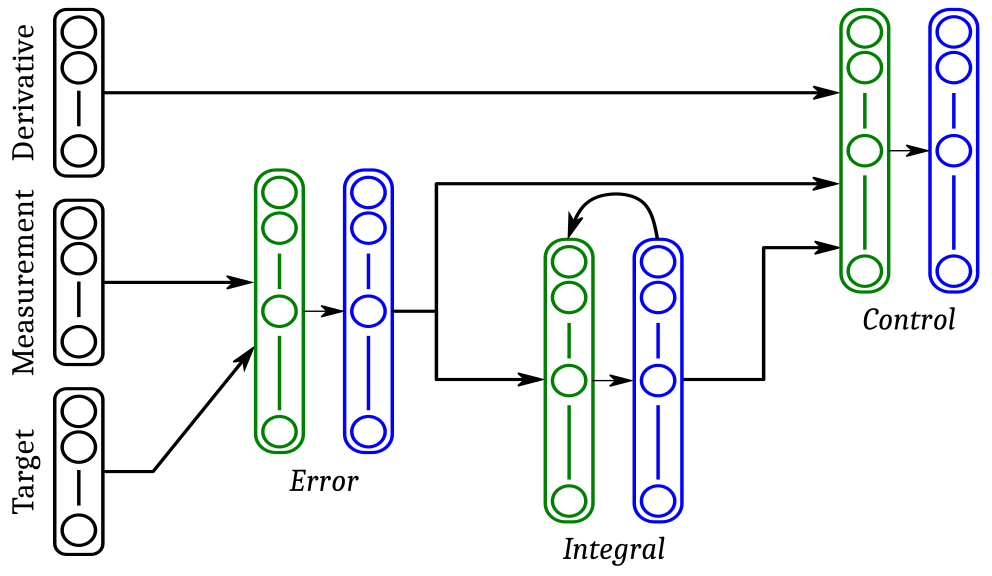}
    \caption{Overall structure of a single PID unit. Green layers stand for aggregate populations of neurons, and blue layers represent the reduce populations (cf. Fig.~\ref{fig:spiking_adder}).}
    \label{fig:snn_full}
\end{figure}

By combining these populations of neurons performing additions and subtractions, the N-PID can be implemented, as shown in Fig.~\ref{fig:snn_full}. Since this design requires three of the previously described neuron populations (cf. Fig~\ref{fig:spiking_adder}), the total amount of neurons necessary for the entire N-PID is $6N + 3$ if all sub-units have the same resolution, and excluding the input neurons ensuring the position encoding of the information.

\subsubsection{Integral wind-up}

Integral wind-up is a well-known problem in PID design~\cite{aastrom2013computer}. By taking the integral over time, error accumulates and the system might experience overshoot. One of two integral wind-up solutions is to bound the size of the integral, so it will not grow too large. Because of the nature of position-coding, this is already inherent to our N-PID. Another solution is to add a decay factor to the integral, "forgetting" errors that lay further in the past. This can be implemented in this N-PID by multiplying the weights of the previous integral signal by a decay factor, that ensures faster decay of the integral term when the set-point is crossed and will be used in our experiments. 

\subsection{Simulation setup}

Both the cascaded PID and the N-PID were tested in simulation. An AscTec Hummingbird quadrotor was simulated in RotorS, a ROS/Gazebo physics simulator that enables quick testing of ROS packages before implementing in a real system~\cite{furrer2016rotors}. State estimation was achieved by utilizing the MSF framework, providing 6-DOF state estimates based on an Extended Kalman Filter (EKF)~\cite{lynen13robust}. The simulator runs on a Dell XPS 17 (Intel i7-10750H 12-core 2.6GHz CPU and 16GB RAM) running Ubuntu 18.04 LTS. The cascaded PID was implemented as a ROS package in Python (\textit{version 3.6}), while the N-PID has been implemented for height control. 

To make sure that the N-PID can run on the Loihi chip, a transformation from floating point weights is necessary. Loihi weights supports 8-bit resolution, with 1 bit reserved for the sign. This means the weights have to be between -256 and 254, with steps of 2. Due to the discretization inherent to this type of network, the first-order derivative suffers largely from jumps in the input bins, especially with lower precision. To counter that, the error derivative is directly fed to the control-layer of the N-PID. 

Besides, it is essential to have an accurate error term in the controller of an unstable non-linear system such as a quadrotor. This controller receives set-points and state estimates and produces thrust and torque commands. These commands are then transformed into rotor velocity commands, which are then sent to the quadrotor model implemented in RotorS. In this simulation, the step responses of the quadrotor to a change in the height set-point are simulated between 1 and 3 meters.

\subsection{Hardware setup}

\begin{figure}
    \centering
    \includegraphics[width=\linewidth]{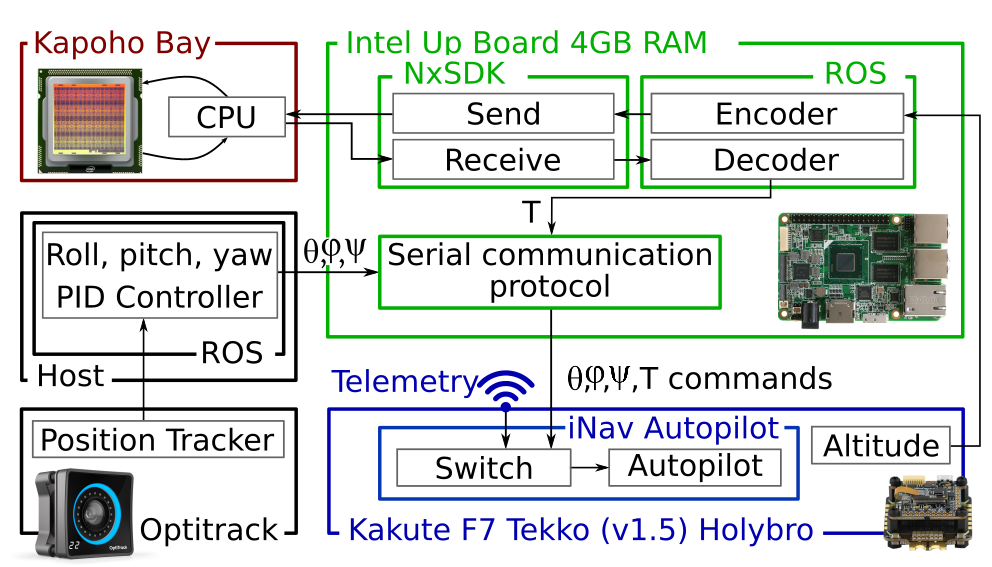}
    \caption{Overview of the hardware architecture of the MAV equipped with the Kapoho Bay. In the context of this study, the neuromorphic PID for height control runs on the Kapoho Bay (thrust $T$), while the roll ($\theta$), pitch ($\phi$) and yaw ($\psi$) controller is done on the host machine which communicates with the quadrotor via a wireless UDP protocol. In this graph, the colored boxes (Up board, Kakute, and Kapoho Bay) indicate indicate that these elements are on-board the drone, while the black boxes (host and OptiTrack) remain outside the drone.}
    \label{fig:hardware}
\end{figure}

The performance of the neuromorphic controller in the real-world has been assessed. The controller was implemented to control the height of a quadrotor. The proposed architecture features a custom made 5-inch MAV equipped with (i) the Kakute F7 flight controller running the iNav 2.6.0 open-source firmware equiped with a TFmini Micro-Lidar for altitude measurements, (ii) the Intel UP board (1.92GHz 64bit Atom processor with 4GB RAM) running Ubuntu 18.04 LTS, and (iii) the Intel Kapoho Bay neuromorphic computing unit incorporating two Loihi chips (Fig.~\ref{fig:hardware}). The entire project has been developed within the ROS framework (Robot Operating System) running onboard the UP board. The communication between the flight controller and the UP board is ensured by serial communication utilizing the MultiWii Serial Protocol (MSP). On the other side, the communication between the UP board and the Kapoho Bay is performed by means of Intel's NxSDK API~\cite{lin2018programming} with ROS-compatible modules, and physically achieved via the USB interface available on the Kapoho Bay. 

The altitude sensor (Micro-LiDAR) provides height measurements sent to the UP board, where the N-PID calculates the desired thrust and sends this information back to the flight controller. The attitude commands are sent from a base station based on the position measurements provided by the Optitrack motion tracking system. 

\section{Results}

\subsection{Simulation results}

The step response of the N-PID controller was first compared to the standard PID in the context of altitude control of a quadrotor for varying altitude set-points ranging from 1.0 to 3.0 meters with a step of 0.5 meters. The N-PID provides the thrust-offset $T$ from hover-thrust. An initial estimate of hover-thrust is obtained by multiplying the weight of the quadrotor with $g=9.81$, further adjusted by hand to account for uncertainties in the model. The gains of both PIDs were tuned by hand ($K_P=0.87$, $T_I=0.17$, $T_D=2.76$). The target and measured value ranges were both chosen within $[0, ..., 4]$ meters and the derivative of the error ranged from -0.5 to 0.5 meters. The range of the output control $T$ was chosen within $r = [-1.25, ..., 1.25]$ (offset for the hover-thrust). 

\begin{figure}
    \centering
    \includegraphics[width=\linewidth]{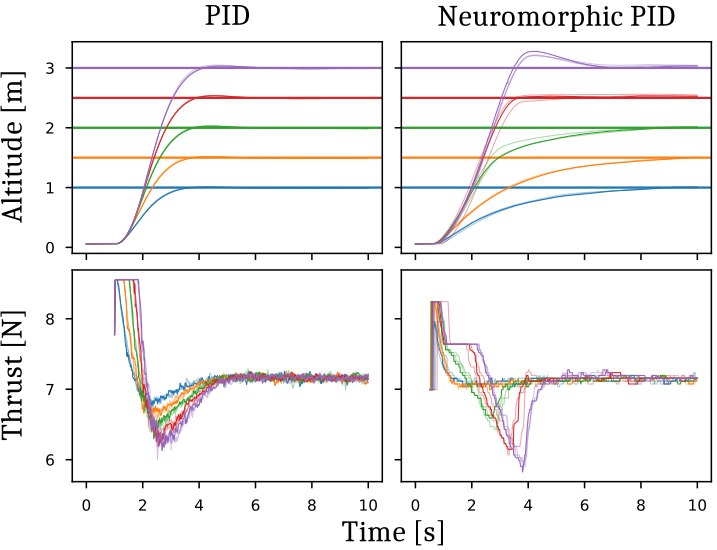}
    \caption{Step response of the N-PID with a population of 151 output neurons, compared to the standard PID for different height set-points (1.0, 1.5, 2.0, 2.5, and 3.0 meters). These results were obtained in simulation.}
    \label{fig:pidvsnpid}
\end{figure}

In Fig.~\ref{fig:pidvsnpid}, we provide the step response of the conventional PID and the N-PID, provided with an overall precision of 0.017N (151 output neurons uniformly distributed). For each altitude set-point we ran 5 distinct simulations. Although both systems manage to reach the target, we notice that the N-PID exhibits a slower dynamic than its conventional counterpart, with higher overshoot for high altitude set-points. Where the conventional PID has a similar settling behaviour for all heights, the N-PID has undershoot at lower heights, but overshoot at higher ones. 

\begin{figure*}
    \centering
    \includegraphics[width=\textwidth]{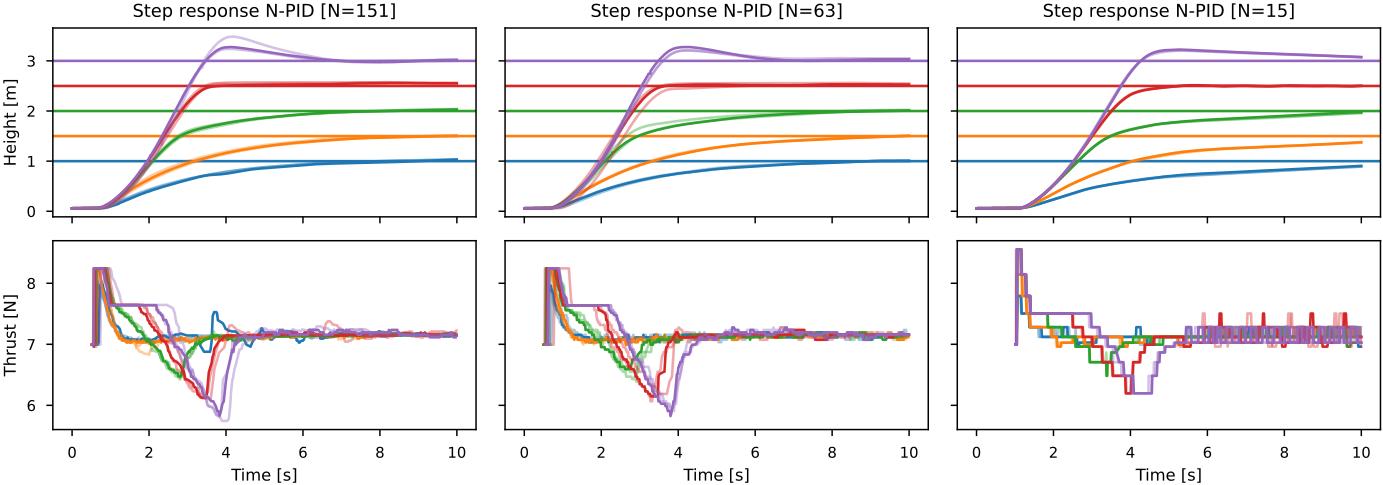}
    \caption{Step responses of the simulated neuromorphic PID (N-PID) applied to height control of the MAV and for varying precision (i.e., 151, 63, and 15 neurons). The target altitudes are 1.0, 1.5, 2.0, 2.5 and 3.0 meters.}
    \label{fig:precision}
\end{figure*}

We further investigated the effect of the control precision (i.e., the number $N$ of output neurons) on the step response of the N-PID. The results are shown in Fig.~\ref{fig:precision} for a number of output neurons $N \in [151, 63, 15]$. For the two first tests ($N=151$ and $N=63$), the population of output neurons obeys a linear distribution, thus ensuring all neurons contribute to the exact same control resolution. Results show that the size of the population of output neurons does not have a significant impact on the overall dynamics of the N-PID.

In case of a population of only 15 neurons, the control range was changed to a quadratic distribution as follows: $\text{sign}(r)\cdot r^2$. This way, the control around hover-thrust is more precise than far from it, hence compensating for the very low resolution that a uniformly distributed population of 15 neurons would have led to (0.17N). In simulation, we observed that the N-PID does not exactly reach the set-point because of the discretization in the target (and measured) position. This means that, because the range of sensitivity of each neuron is quite large, the error has a higher chance to be equal to zero while the target altitude has not been reached yet. Nevertheless, the results obtained with the N-PID endowed with a population of only 15 neurons to represent the control command with a quadratic sensitivity distribution are promising and suggest that this setup should be reliable enough to be further applied on the MAV.

\begin{figure}
    \centering
    \includegraphics[width=0.9\linewidth]{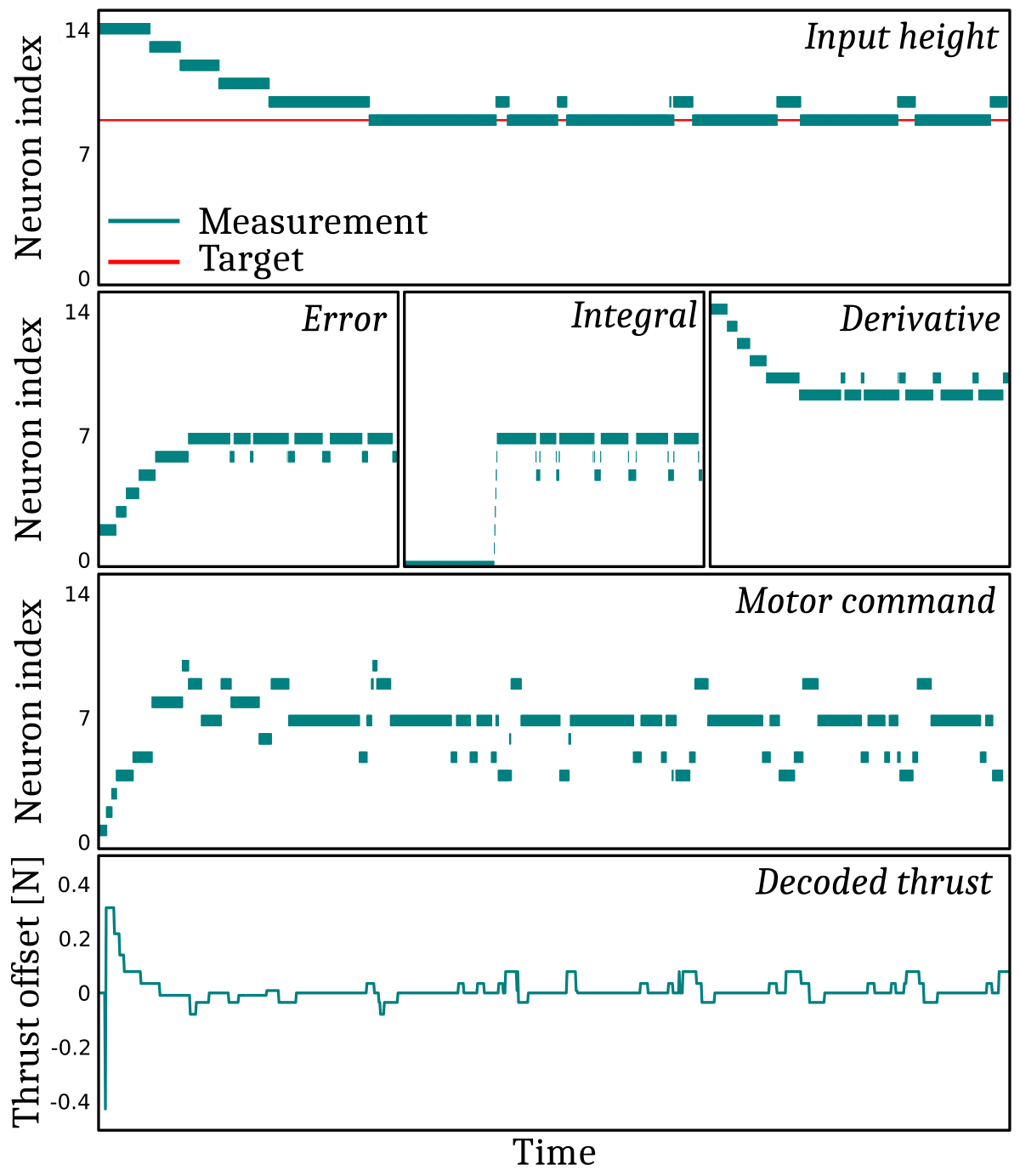}
    \caption{Overview of the spiking activity on-board the Loihi chip running the N-PID height controller in simulation (offline data). In this case, the target altitude is 1.5 meters. The thrust offset corresponds to the offset to be added to the hovering thrust command.}
    \label{fig:offline_loihi}
\end{figure}

As the final goal is to have the N-PID running on-board the Loihi chip mounted on the MAV, a thorough comparison of the spiking activity of the network has been performed to compare the simulated N-PID to the network implemented on the Loihi. This analysis resulted in a almost perfect match between the two setups, thus ensuring that the switch from conventional to neuromorphic hardware is not increasing the reality gap. An overview of the spiking activity recorded on the Loihi is provided in Fig.~\ref{fig:offline_loihi} for a target altitude of 1.5 meters. In order to simplify the visualization, only the output layers of sub-units (error, integral, motor command) are shown. The sparsity of the spiking activity contributes to the parsimony of the proposed N-PID. The average execution time per time step of the entire network has been assessed on the Loihi, showing a total of $2 \mu s$ per time-step, thus resulting in an average update frequency of 500kHz, which is far beyond what we need for online applications.

\subsection{Real-world results}

\begin{figure}
    \centering
    \includegraphics[width=\linewidth]{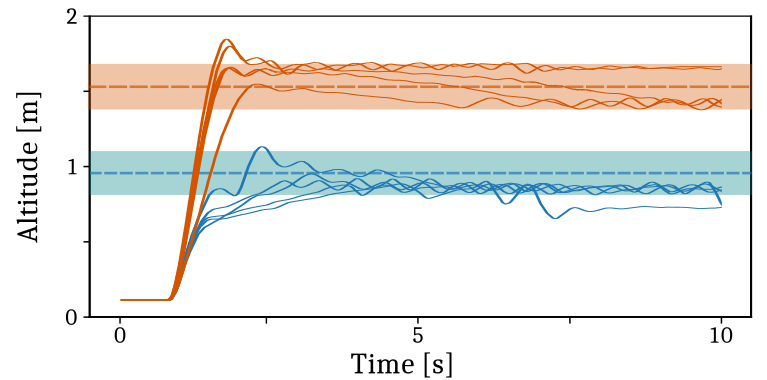}
    \caption{Real-world tests for take-off altitude control with neuromorphic hardware for $N=15$. Different set-points are shown in different colors. The shaded bands around the set-point represent the bin around the set-point altitude}
    \label{fig:online_loihi_full}
\end{figure}

In the following, we implemented the N-PID for altitude control on the Kapoho Bay neuromorphic device mounted on-board a MAV. Based on the results obtained in simulation, we decided to test the small network with a population of 15 output neurons with a quadratic distribution of the control precision. Therefore, the N-PID features only 93 neurons ($6N + 3$), plus 45 input neurons for the encoding of the altitude measurement and target, along with the derivative of the error. Two different altitude set-points were tested (1.0m and 1.5m) over 5 trials each. The altitude ground truth over time was given by the OptiTrack motion capture system, and the results are shown in Fig.~\ref{fig:online_loihi_full}. The shaded colored areas show the target bins, caused by the precision of the altitude discretization. Inside the shaded area, the error between the measured and target altitude is zero, which means the MAV is flying at hover-thrust. For some flights it can be seen that the drone is not able to maintain altitude. On the real drone, the hover thrust PWM level depends largely on the battery level. Even though it is visible that the integrator is able to compensate for a voltage drop over multiple runs, the integral output will be saturated and the MAV will lose altitude with a low battery. During the tests, the control loop update frequency of the system was set to 70Hz to be consistent with the height sensor. 

\begin{figure}
    \centering
    \includegraphics[width=0.95\linewidth]{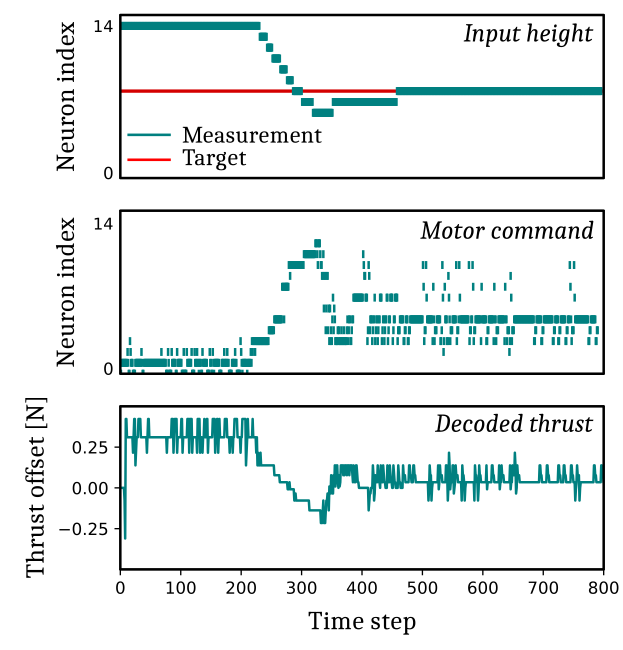}
    \caption{Overview of the Loihi input/output spike activity during an online test with the MAV (target altitude: 2.0m), as well as the decoded thrust offset (in Newtons). The controllers generates a slight overshoot for high targets, but quickly converges to the desired set-point.}
    \label{fig:online_loihi_sample}
\end{figure}

In Fig.~\ref{fig:online_loihi_sample} the I/O spiking activity of the N-PID running on the neuromorphic chip is shown for one of the tests performed with the MAV. Compared to the simulation results, there is more noise around the thrust command. This is caused by the derivative that is based on the measured altitude and has a resolution of 1 cm.

\section{Conclusion}

We presented a neuromorphic implementation of the PID to control the altitude of a MAV equipped with the Loihi neuromorphic processor. Using a very low number of neurons, the network demonstrated its capability of achieving robust and stable control, with the potential of reaching extremely high control loop frequencies. The proposed neural network does not require any training as its circuitry inherently captures the desired dynamics -- only the weights of the synapses need to be tuned to fit with the aerodynamic properties of the MAV. In practice, the tuning of the input, error and control ranges, can be difficult for unstable systems with a low amount of neurons. Here, using non-linear distributions for these ranges has proved to mimic the response of a controller with a larger amount of neurons and is very promising for attitude control.

In both Stagsted et al.~\cite{stagsted2020towards} and Zaidel et al.~\cite{zaidel2021neuropid}, it was observed that implementing a proper derivative action on neuromorphic hardware is difficult and remains an open problem. Future improvements of this algorithms include a fully neuromorphic implementation of the N-PID for controlling all degrees of freedom of the MAV, including attitude commands. This requires multiple PIDs to run in parallel and series on the neuromorphic hardware.

Having a neuromorphic controller like a PID that is easily implementable on available hardware contributes to closing the neuromorphic loop for control in robotics. These controllers can be easily implemented in pipelines making use of event-based algorithms.

Lastly, the very high execution frequency of the N-PID on the Loihi chip (around 500kHz) strengthens the fact that conventional hardware (including communication protocols) must be improved to meet the promise of neuromorphic computing to target very fast control in tasks such as drone racing and aggressive maneuvers with MAVs.




\section*{Supplementary information}

The codes and data collected during this study are publicly available online at the following address: \url{https://github.com/tudelft/neuro_pid}. An additional video recording of the tests with the quadrotor is provided with the paper.



\bibliographystyle{IEEEtran}
\bibliography{biblio}

\end{document}